\documentclass[letterpaper]{article} 
\usepackage{aaai23}  
\usepackage{times}  
\usepackage{helvet}  
\usepackage{courier}  
\usepackage[hyphens]{url}  
\usepackage{graphicx} 
\usepackage{natbib}  
\usepackage{caption} 
\usepackage{algorithm}
\usepackage{algorithmic}

\usepackage{amssymb}
\usepackage{amsmath}
\usepackage{multirow}

\usepackage{newfloat}
\usepackage{listings}
\DeclareCaptionStyle{ruled}{labelfont=normalfont,labelsep=colon,strut=off} 
\floatstyle{ruled}
\newfloat{listing}{tb}{lst}{}
\floatname{listing}{Listing}
\title{Dual Memory Units with Uncertainty Regulation for Weakly Supervised \\Video Anomaly Detection}
\author{
    Hang Zhou,
    Junqing Yu\thanks{indicates co-corresponding author.},
    Wei Yang \footnotemark[1]
}
\affiliations{
    Huazhong University of Science and Technology, Wuhan, China\\
    \{henrryzh, yjqing, weiyangcs\}@hust.edu.cn
}

\usepackage{bibentry}

\begin{document}

\maketitle

\begin{abstract}
    Learning discriminative features for effectively separating abnormal events from normality is crucial for weakly supervised video anomaly detection (WS-VAD) tasks. Existing approaches, both video and segment-level label oriented, mainly focus on extracting representations for anomaly data while neglecting the implication of normal data. We observe that such a scheme is sub-optimal, i.e., for better distinguishing anomaly one needs to understand what is a normal state, and may yield a higher false alarm rate.
    To address this issue, we propose an Uncertainty Regulated Dual Memory Units (UR-DMU) model to learn both the representations of normal data and discriminative features of abnormal data.
    To be specific, inspired by the traditional global and local structure on graph convolutional networks, we introduce a Global and Local Multi-Head Self Attention (GL-MHSA) module for the Transformer network to obtain more expressive embeddings for capturing associations in videos.
    Then, we use two memory banks, one additional abnormal memory for tackling hard samples, to store and separate abnormal and normal prototypes and maximize the margins between the two representations.
    Finally, we propose an uncertainty learning scheme to learn the normal data latent space, that is robust to noise from camera switching, object changing, scene transforming, etc. 
    Extensive experiments on XD-Violence and UCF-Crime datasets demonstrate that our method outperforms the state-of-the-art methods by a sizable margin.
\end{abstract}

\section{Introduction}
\label{sec:intro}
Video anomaly detection (VAD) aims to identify the time window of abnormal events in untrimmed videos~\cite{Lu,Pred,MIL,XD}, and has beneficial applications in the context of intelligent video surveillance. 
Yet collecting fine-grained anomaly videos for training are arduous, weakly supervised approaches, which assign one label, i.e., normal or abnormal, for each panoptic video, have been gaining popularity and achieving state-of-the-art performances~\cite{GCN,CLAWS,RTFM}. 
Following weakly supervised setup, multiple instances learning (MIL) is one of the most effective strategies~\cite{Deepmil,AttMIL}. MIL-based anomaly detection methods consider each video as a bag, where abnormal videos are defined as positive and normal videos as negative. Meanwhile, anomaly snippets are presented as positive instances in positive bags, and normal snippets are negative instances. MIL-based methods then learn the instance-level labels from video bags by selecting top-K instances as abnormal according to discriminative representations~\cite{IBL,MA,Wustad,CRFD}.

\begin{figure}[t]
    \centering
    \includegraphics[width=0.95\columnwidth]{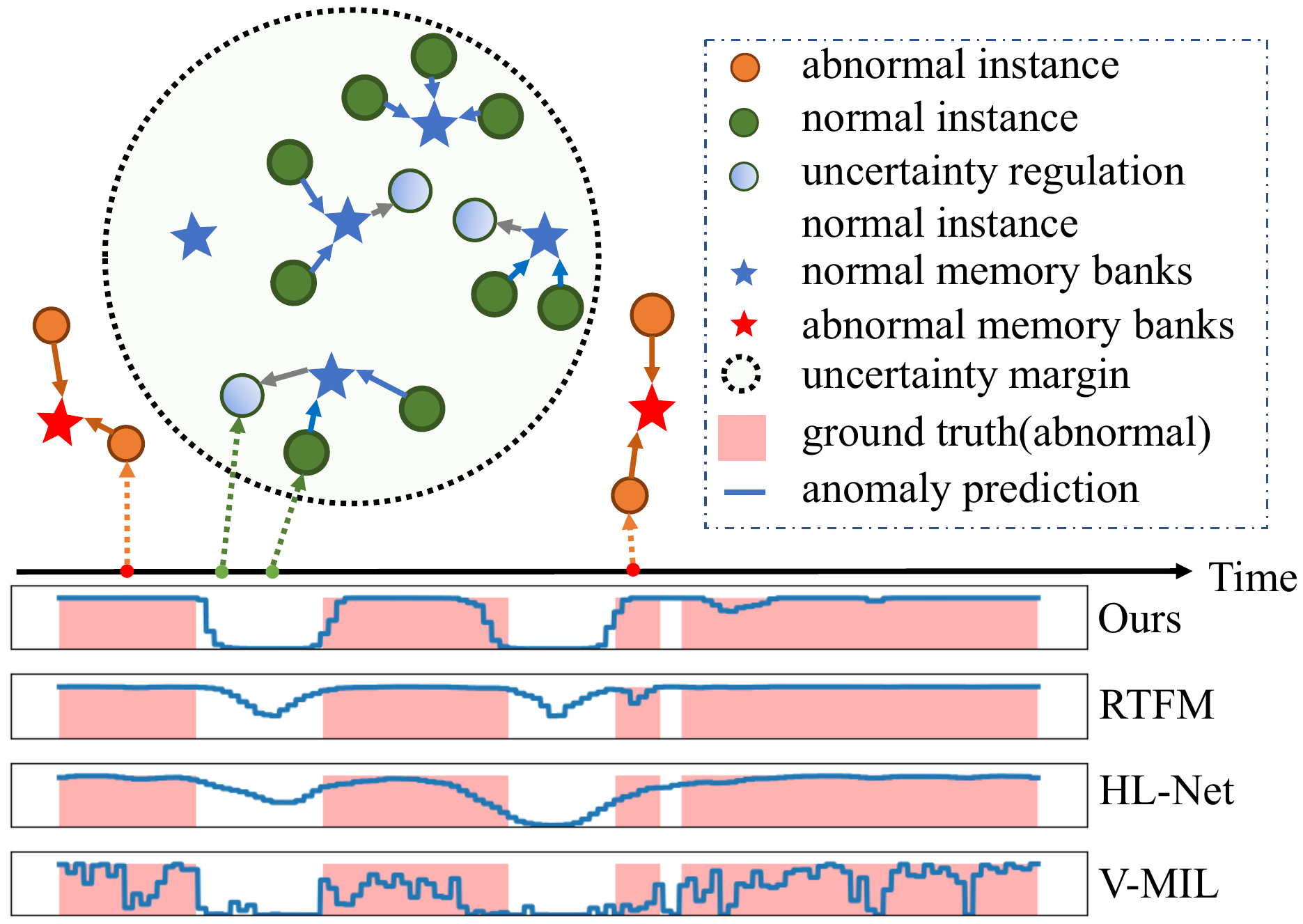} 
    \caption{We use two memory banks to store normal and anomaly instances respectively,
    both memory banks attract close instances. The normal space is regulated to be a Gaussian distribution, while no additional regulation applied for anomaly feature space. 
    Such a strategy can tackle the hard samples better, the bottom curve shows prediction results of a part of the video ``Black.Hawk.Down''.
    }
    \label{teaser}
\end{figure}

Notably, MIST~\cite{MIST} develops a two-stage self-training procedure by generating pseudo labels using MIL approaches and emphasizing anomalous regions with a self-guided attention boosted feature encoder. RTFM~\cite{RTFM} uses a feature magnitude learning function to recognise the positive instances, effectively improving the robustness of the MIL approach. Nevertheless, existing weakly supervised approaches focus on learning discriminative features for anomalous instances, while neglecting the implication of normal data, and hence yield high false alarm rate. 
As shown in the bottom of Figure \ref{teaser}, RTFM and HL-Net both lead to alarm falsely.

To address the problems and better distinguish anomaly, we propose a dual memory units module with uncertainty regulation to store normal and abnormal prototypes respectively and learn their distinctive distributions, as shown in the top of Figure \ref{teaser}.
Previous approaches use only one memory for normality, while we observe such strategy can not handle hard samples well, e.g., the action of running exists in both normal exercising videos and criminal arresting videos, and hence yield a high false alarm rate. Dual memory units consider more peripheral factors compared to a single memory, and may reduce false alarms. 
Specifically, we encourage the anomaly memory bank to intake more anomaly information from videos annotated as anomaly, while the normal memory bank to obtain normal data pattern from both normal and abnormal videos.
Following RTFM's paradigm which indicates that normal features are in low magnitude while anomaly features are on the contrary, we observe that normal features exhibit fluctuations due to noises coming from camera switching, subject changing, etc., and further format low magnitude normal features with Gaussian distribution. We adopt a normal data uncertainty learning (NUL) scheme to learn both the mean feature and uncertainty variance, and consequentially generate a latent normal space. We can segregate the normality and anomaly according to the latent space and reduce the false alarm rate with noise perturbations. 
Moreover, inspired by the traditional global and local structure on graph convolutional networks, we introduce a Global and Local Multi-Head Self Attention (GL-MHSA) module for the Transformer network to obtain more expressive embeddings for capturing associations in videos.

In summary, the main contributions of this paper are as follows:
\begin{itemize}
    \item We design a Global and Local Multi-Head Self Attention model to capture different associations of anomaly videos for enhancing the expressive embeddings.
    \item We propose the dual memory units, which use both normal and abnormal memory banks for better distinguishing patterns of normality and anomaly. We also introduce a dual memory separation loss to learn more robust and discriminative features.
    \item We propose a normal data uncertainty learning scheme to enhance the normal pattern representation ability and better separate the anomaly from latent normality.
  
\end{itemize}

\section{Related Work}
\label{sec:relat}

\textbf{Weakly Supervised Video Anomaly Detection.}
VAD is an important task in computer vision~\cite{dlsurvey}, and mainstream methods either adopt an unsupervised learning strategy with only normal data or weakly supervised learning setup with video-level annotated data. Unsupervised methods with only normal data can not well handle complex environments, and stimulated by progress in weakly supervised learning in video action localization~\cite{STPN,Basnet,UM,ACG}, WS-VAD gradually gains popularity.
\cite{MIL} introduce a large real-world surveillance dataset and proposed a vanilla MIL (V-MIL) framework with ranking loss. \cite{GCN} consider the normal label to contain noise and apply pseudo label generation method by graph convolutional network to tackle noise specifically. 
Recent work mostly adopt the MIL training strategy. 
Most improvements expand the receptive field and enhance the instance embedding \cite{CRFD}, e.g., RTFM proposes Multi-scale Temporal Network (MTN) to capture the long and short-range temporal dependencies.
Other approaches focus on multi-modal setup, for instance \cite{ACF,Pang} use both audio and video information for VAD. 

Self-training strategy generate labels from coarse to fine level \cite{crst,stdor} and has been explored in semi-supervised learning.
In VADs, \cite{stdor} present an end-to-end and self-training method for anomaly regression without manually labeled training data.
\cite{MIST} propose a multiple instance pseudo label generator and self-guided attention boosted encoder to refine discriminative representations.
\cite{MSL} apply multi-sequence learning with Transformer to refining abnormal scores through self-training method.
%
Both types of methods consider normal snippets are noisy and lack normal data learning, while learning anomalous patterns excessively without normal contrast suffers from overfitting.
In contrast, we adopt the MIL setup and use memory banks to store the normal and abnormal patterns. Moreover, we apply normal data uncertainty regulation to learn the latent normal space and enlarge the distance between normality and anomaly.

\textbf{Memory Networks.}
Memory network is an efficient neural network to store different prototypes and has been applied in various tasks \cite{maganad,AUMN}.
\cite{MemAE} propose a memory augmented deep autoencoder to suppress the reconstruction of anomaly.
\cite{MNAD} follows the MemAE framework and presents a new readable and updatable memory scheme, where memory banks also update at test time.
\cite{HF2} design a multi-level memory module in an autoencoder with skip connections to memorize normal patterns for optical flow reconstruction.
All the above methods use memory banks to learn from normality. In this work, We introduce dual memory units to store both normal and abnormal prototypes, and we design a dual loss to guide memory banks in prototype learning.

\begin{figure*}[t]
    \centering
    \includegraphics[width=0.9\textwidth]{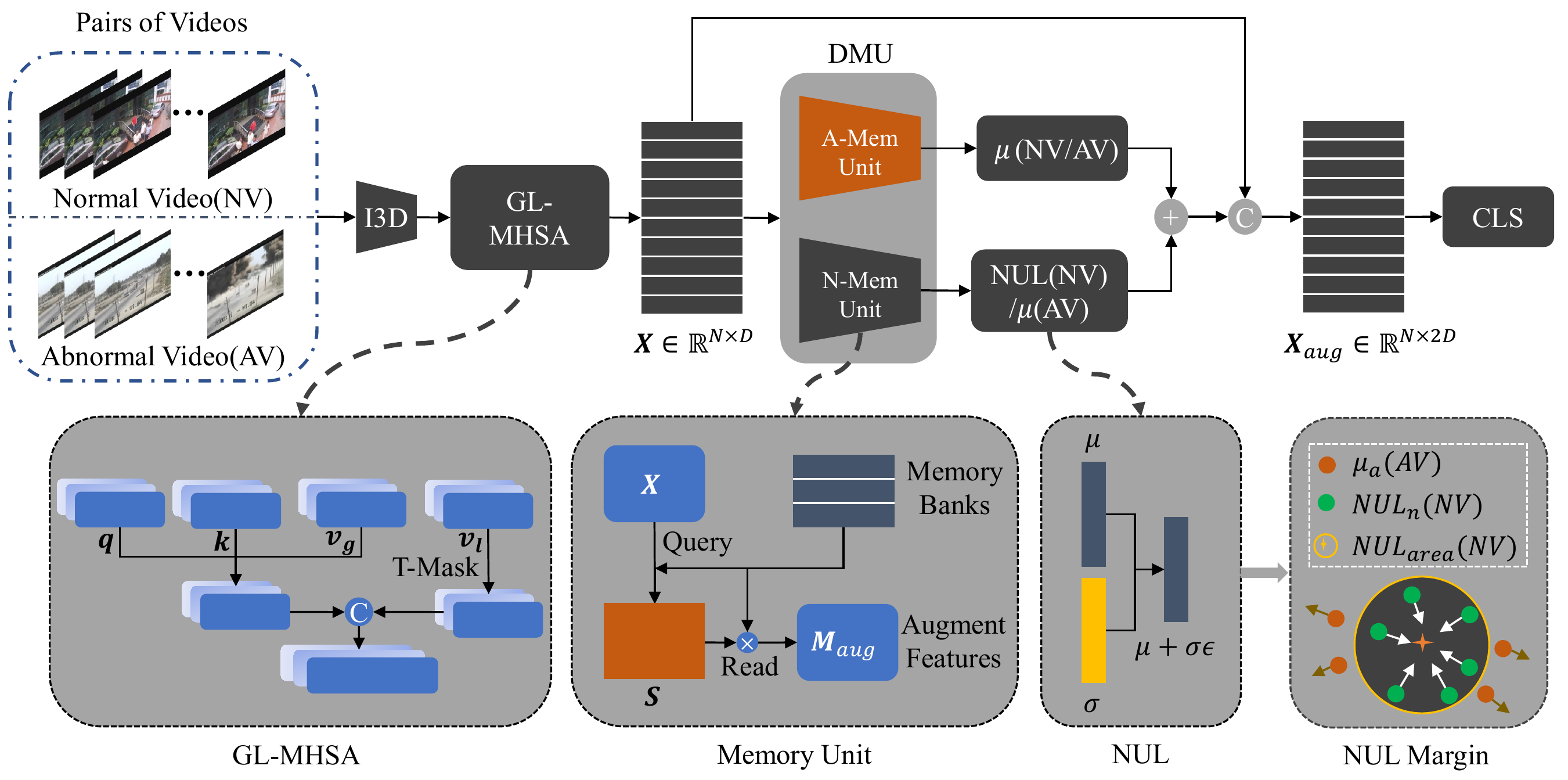} 
    \caption{The framework of our UR-DMU model consists of three parts:
    global and local feature learning (GL-MHSA), dual memory units (DMU), and normal data uncertainty learning (NUL). GL-MHSA extracts better expressive embeddings, DMU stores both normal and abnormal patterns for discrimination, and NUL constrains the normal data as a Gaussian distribution to handle uncertainty perturbation.
    }
    \label{fig1}
\end{figure*}

\textbf{Data Uncertainty Learning.}
Data uncertainty learning captures the data variance due to noise and randomness~\cite{DUL,UMOOD,UM}.
\cite{DUL} propose the data uncertainty learning scheme to learn the feature embedding and its uncertainty in face recognition.
In video temporal action localization, \cite{UM} formulate the background frames as out-of-distribution (OOD) samples and proposed an uncertainty model to learn the uncertainty. 
Different from \cite{UM}, we presume that the anomaly data is OOD, and the normal data follows a Gaussian distribution. The latent space of the normal uncertainty pattern is effective for unknown normal samples. Meanwhile, it also helps the model to separate normality and anomaly.

\section{Method}
\label{sec:app}
Here we introduce the Uncertainty Regulated Dual Memory Units (UR-DMU) method to store both normal and abnormal prototypes and enforce additional prior distributions in details. As illustrated in Figure \ref{fig1}, UR-DMU consists of three modules: the Global and Local Multi-Head Self Attention (GL-MHSA) for feature learning, the Dual Memory Units(DMU) for storing prototypes, and the Normal Data Uncertainty Learning (NUL) for normality latent embedding learning.
In the following sections, we briefly introduce the formations of our problem first, and then describe the above three modules respectively, and finally show the training and testing process.

\subsection{Problem Formation}
Following the MIL setup for WS-VAD tasks, we define normal videos as $\{ v_i^n\} _{i=1}^N$ and abnormal videos as $\{ v_i^a\} _{i=1}^N$. Each anomaly video is a bag $Y^a=1$ with at least one abnormal clip, similarly a normal video is marked as $Y^n=0$ with all normal clips.
Our goal is to score each $v_i$ according to $Y_i=f(v_i)$ with the model $f(\cdot)$.
It is a common practice to extract snippet features using pretrained backbones~\cite{c3d,i3d,vswin} and conduct future processes on extracted features, in this paper we choose I3D model pre-trained on Kinetics dataset as our backbone, and extract RGB features of consecutive 16-frames~\cite{XD}. The extracted video features are denoted as $ \textbf{X}_{i3d} \in \mathbb{R}^{N \times F}$ , where $N$ represent the number of snippets, and $F$ is the dimension of extracted features.

\subsection{Global and Local Feature Learning}
Learning long and short temporal dependencies of anomalous features is crucial for VAD~\cite{nolocal,stg}. Traditional approaches adopt the graph convolution network for extracting global and local associations~\cite{XD}. Recently, the transformer structure shows its superiority in handling long-range associations between temporal signals~\cite{trans,vit,swin,focal}.
For better capturing long and short temporal dependencies simultaneously, we borrow the idea of global and local structure in GCN and develop the Global and Local Multi-Head Self Attention (GL-MHSA) for transformer. 
Different from the conventional MHSA, we add an extra encoder layer of $\textbf{V}_l=\textbf{X}_{i3d} \textbf{W}_l$ to learn local features with a temporal mask:

\begin{equation}
    \textbf{T}_{m}(i,j)=-\frac{|i-j|}{e^\tau}.
\end{equation}
Then we normalize the mask with $\text{softmax}(\textbf{T}_{m})$, where $\tau$ is the sensitivity factor for balancing the local distance. 
While computing $Q, K, V$, we obtain the global and local feature $\textbf{X}\in \mathbb{R}^{N \times D}$ from Multilayer Perceptron (MLP) and LayerNorm (LN), which can be formulated as follows:
\begin{equation}
    \begin{aligned}
        \textbf{X}&=\text{MH}([ \text{softmax}(\frac{\textbf{Q}\textbf{K}^T}{\sqrt{D}})\textbf{V}_g; \text{softmax}(\textbf{T}_{m})\textbf{V}_L ]),\\
        \textbf{X}&=\text{LN}(\text{MLP}(\text{LN}(\textbf{X}))+\textbf{X}).
    \end{aligned}
\end{equation}

\subsection{Dual Memory Units}
Previous approaches use memory units to store prototypes of normal data for better scoring. We observe that a single unit is insufficient for distinguishing hard samples, and we use learnable normal and abnormal memory banks to store corresponding templates $\textbf{M}\in \mathbb{R}^{M \times D}$,
where $M$ is the memory bank number, $D$ is the output feature dimension.
We apply query and read operations to encourage correct template learning, and by querying memory banks $\textbf{M}$ we can assess the similarities between features:
\begin{equation}
    \begin{aligned}
        \textbf{S}&=\sigma (\frac{\textbf{X} \textbf{M}^T}{\sqrt{D}}),\hspace{1em} \textbf{M}_{aug}=\text{S}\textbf{M}, 
    \end{aligned}
\end{equation}
where $\sigma$ is sigmoid activation, and $\textbf{S} \in \mathbb{R}^{N \times M}$ is the query score. Then the memory augmentation feature generated by a read operation is represented as $\textbf{M}_{aug}$. $\textbf{S}$ indicates whether snippets are similar to memory banks. 
We apply topK selection along the second dimension to determine which snippets are most similar to memory banks:
\begin{equation}
    \begin{aligned}
        \textbf{S}_{k;ind},\textbf{S}_{k;sc}=\text{topK}(\textbf{S},K),\hspace{0.5em} \textbf{S}_{k;i}&=\frac{\sum_{j=1}^{K} S_{k;sc}(i,j)}{K}.\\
    \end{aligned}
    \label{topk}
\end{equation}
In equation (\ref{topk}), $K\ll M$, $\textbf{S}_{k;ind}$ means the indices of proposed K snippets to be stored, and $\textbf{S}_{k;sc} \in \mathbb{R}^{N \times K}$ are their query scores.
$\textbf{S}_{k} \in \mathbb{R}^{N \times 1}$ indicates the highest scores of memory matching. 

\begin{figure}
    \centering
    \includegraphics[width=0.95\columnwidth]{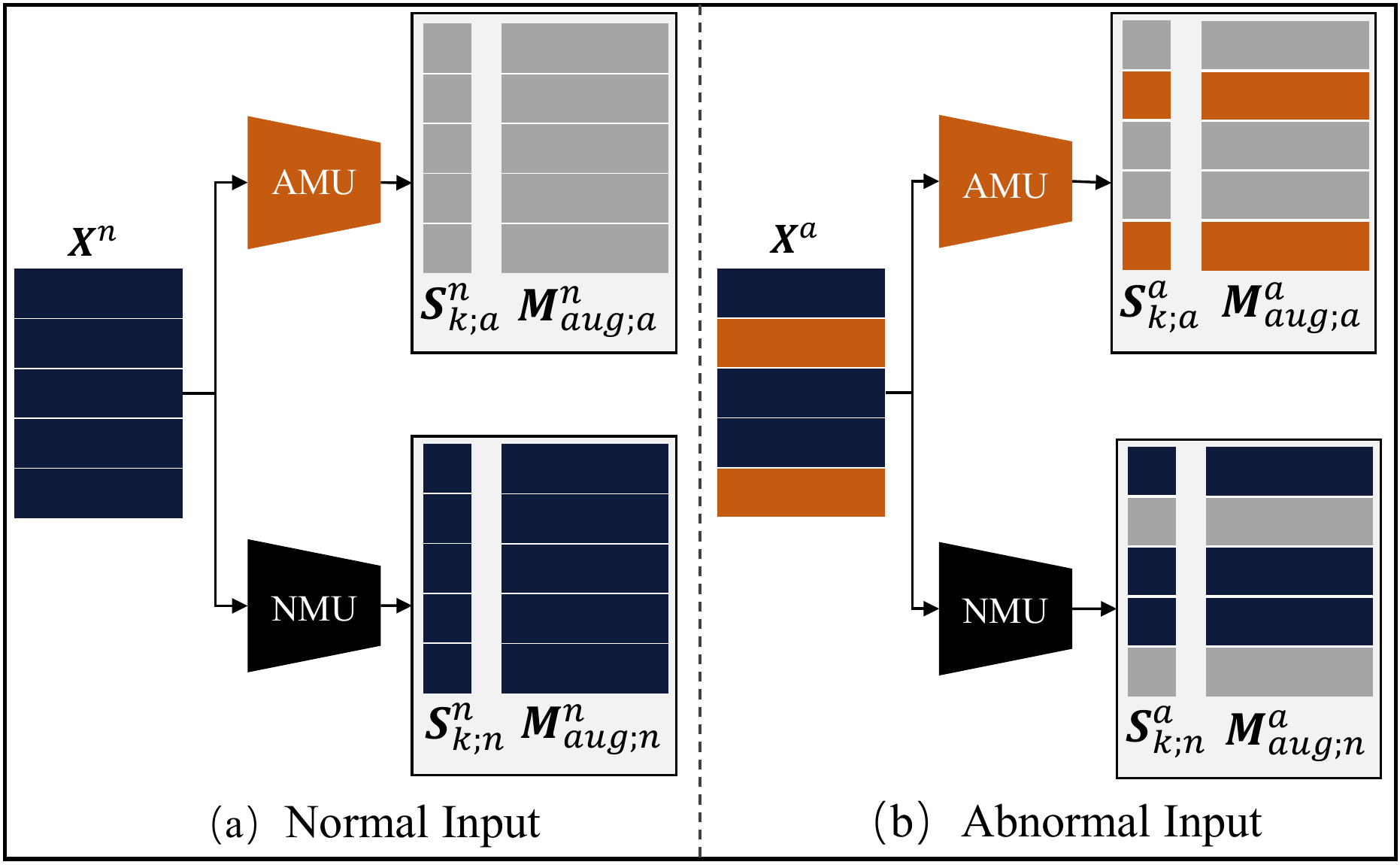} 
    \caption{The dual memory units store normal and anomaly instances in different memories. (a) shows that normal video feature goes through dual memory units and obtains relevant output scores and augment features.
    (b) is the case with abnormal video feature input. 
    }

    \label{fig2}
\end{figure}

Figure \ref{fig2} shows different input features across the dual memory units. 
While sending normal video embedding $\textbf{X}^n \in \mathbb{R}^{N \times D}$ into the abnormal memory banks, $\textbf{S}_{k;a}^n$ and $\textbf{M}_{aug;a}^n$ are generated by querying and reading operations. Similarly, we can also obtain $\textbf{S}_{k;n}^n$ and $\textbf{M}_{aug;n}^n$ as video embedding goes through normal memory banks.
The anomaly query scores $\textbf{S}_{k;a}^n$ are constrained to $\textbf{0} \in \mathbb{R}^N$ because of the normal input.
And the normal query score $\textbf{S}_{k;n}^n$ should be $\textbf{1} \in \mathbb{R}^N$.

Likewise, when the abnormal video embedding input to abnormal memory banks, we obtain $\textbf{S}_{k;a}^a$ and $\textbf{M}_{aug;a}^a$. 
Because of the sparseness nature of the anomaly, only several snippets will be activated by anomaly memory banks.
We can also obtain $\textbf{S}_{k;n}^a$ and $\textbf{M}_{aug;n}^a$ from normal memory banks.
Different from normal input, the anomaly query score $\textbf{S}_{k;a}^n$ is not all $\textbf{1}$ because of the weak label of anomaly. 
Same with MIL, there is at least one element equals $1$ in $\textbf{S}_{k;a}^a$ and $\textbf{S}_{k;n}^a$.

To train the dual memory units, we define a dual memory loss consists of four binary cross-entropy (BCE) losses:
\begin{equation}
    \begin{aligned}
        L_{dm}& =  \text{BCE}(\textbf{S}_{k;n}^n,\textbf{y}_n^n)+\text{BCE}(\textbf{S}_{k;a}^n,\textbf{y}_a^n) \\
         & +\text{BCE}(S_{k;n;k}^a,y_n^a)+\text{BCE}(S_{k;a;k}^a,y_a^a),
    \end{aligned}
\end{equation}
where $\textbf{y}_n^n=\textbf{1} \in \mathbb{R}^N$ and $\textbf{y}_a^n=\textbf{0} \in \mathbb{R}^N$. 
$S_{k;n;k}^a,S_{k;a;k}^a \in \mathbb{R}$ are means of $\textbf{S}_{k;n}^a,\textbf{S}_{k;a}^a$ topK result along the first dimension.
The labels $y_n^a,y_a^a$ are equal to $1$.
The first two terms of $L_{dm}$ limit the normal memory unit to learn normal patterns, and the last two terms ensure the abnormal pattern learning for abnormal memory unit.
With $L_{dm}$, memory banks store homologous patterns.


With the query scores $\textbf{S}_{k;n}^n$, $\textbf{S}_{k;a}^n$ and $\textbf{S}_{k;a}^a$, we separate the abnormal and normal feature embedding by triplet loss.
\begin{equation}
    \begin{aligned}
        L_{trip}&=\text{Triplet}(f_{a},f_{p},f_{n}),\\
        f_a&=\text{topK}(\textbf{S}_{k;n}^n;\textbf{X}^n),\\
        f_p&=\text{topK}(\textbf{S}_{k;n}^a;\textbf{X}^a),\\
        f_n&=\text{topK}(\textbf{S}_{k;a}^a;\textbf{X}^a),
    \end{aligned}
\end{equation}
where $f_a$ is the average feature of the interest snippets from $\textbf{X}^n$, which indicates the primary part of normal video.  The snippets are selected by the topK index of $\textbf{S}_{k;n}^n$. Similarly, $f_p$ means the normal part in an abnormal video, and $f_n$ means the abnormal part in the abnormal video.

\subsection{Normal Date Uncertainty Learning}
As pointed out by \cite{RTFM,UM}, the magnitude of the normal feature is lower than that of anomalies, i.e., $\mathbb{E}[\| x^a \Vert_2 ]\geqslant \mathbb{E}[\| x^n \Vert_2 ]$, $x^a$ and $x^n$ are the snippet features of anomaly and normality respectively.
Presumably, magnitude feature learning is also appropriate for the augmentation features $\textbf{M}_{aug}$ of memory output.
Hence, the magnitude separability of augmentation features between abnormal and normal is defined as $d(a,n)=\| m^a \Vert_2 - \| m^n \Vert_2 $.
Moreover, effective representation learning of normality should be able to capture both the core and cover variance introduced by noise. We propose to learn the distribution of normality, and then anomaly is out of this distribution (OOD)~\cite{adood}. This idea has been extensively explored in unsupervised approaches, but rarely used in WS-VAD.
Similar to unsupervised anomaly detection methods, we propose to use the Gaussian distribution to constraint the latent normal representation $z^n$, i.e.:
\begin{equation}
    p(z_i^n|x_{i}^n)= \mathcal{N} (z_i^n;\mu_i;\sigma_i^2I),
\end{equation}
where $\mu$ is the normal feature's mean, and $\sigma$ is the variance (uncertainty).
Both of them can be predicted by an encoder network.
To ensure the normal data sampling operation is differentiable, 
we follow \cite{DUL} and use the re-parameterization to tackle the latent with noise.
The DUL features $z^n$ is generated by random noise $\varepsilon \sim \mathcal{N}(0,1)$:
\begin{equation}
    z_i^n=\mu_i^n+\sigma_i^n\varepsilon, \\
    \label{nul}
\end{equation}
Equation (\ref{nul}) represents the latent space of normal data. 
However, the sample $z$ is impressionable to $\sigma$ when $\sigma$ is too large, and it may be far away from the real features $\mu$.
To suppress the instability of $\sigma$,
a Kullback-Leibler diversity(KLD) regularization term is necessary:
\begin{equation}
    \begin{aligned}
        L_{kl}&=\text{KL}(\mathcal{N}(z^n|\mu^n,\sigma^n)|| \mathcal{N}(0,1))\\    
        &=-\frac{1}{2D}\sum_{i=1}^{D}(1+\text{log}(\sigma_i^n)^2-(\mu_i^n)^2-(\sigma_i^n)^2),    
    \end{aligned}
\end{equation}
  
In testing stage, calculation of the variance can be skipped, and we only need the mean to generate robust prediction.
Furthermore, as mentioned, anomalies are OOD samples, and we need to enlarge the margin between anomalies and normalities for robustness. 
The anomaly features are also fed to the mean-encoder to get the embedding $\mu^a$ in the same space.
We use the magnitude distance loss to separate them in the latent normal space:
\begin{equation}
    L_{dis}=\max(0,d-(\left\lVert \mu_k^a\right\rVert_2^2 -\left\lVert z_k^n\right\rVert_2^2)),
    \label{dis}
\end{equation}
where $\mu_k^a$ are the average features of proposed snippets from $\textbf{M}_{aug}$. 
The snippets are selected by topK index operation from $\textbf{S}_{k;a}^a$.
Same with the above operation, $z_k^n$ is obtained from $\textbf{S}_{k;n}^n$ and $\textbf{M}_{aug}$.

Finally, we fuse the DUL features with $\textbf{X}$ and feed it to the classifier. 
\subsection{Network training and Testing}
\textbf{Training}: Pairs of videos with the same amount of normal and abnormal videos are divided into a batch and inputted to our model.
Then we generate one score for each snippet and use the BCE loss to discriminate the normality and anomaly with five auxiliary losses:
\begin{equation}
    \begin{aligned}
        L&=L_{cls}+\lambda_1L_{dm}+\lambda_2L_{trip}+\lambda_3L_{kl}+\lambda_4L_{dis},
    \end{aligned}
\end{equation}
where $L_{cls}$ is BCE loss with the supervision of video-level labels, 
$\lambda_1,\lambda_2,\lambda_3,\lambda_4$ are hyperparameters.

\noindent \textbf{Testing}: We feed testing videos to our UR-DMU network. In the DUL module, we only use the mean-encoder network to obtain feature embeddings.
Finally, the snippet labels are assigned to frame labels.

\section{Experiments}
\label{sec:exp}
\subsection{Datasets and Evaluation Metrics}
\textbf{Dataset.} Similar to previous works, we conduct experiments on two large video anomaly detection datasets: UCF-Crime \cite{MIL}, and XD-Violence \cite{XD}.

UCF-Crime is a large-scale surveillance dataset that includes 1900 videos with 13 types of anomalous events, e.g., explosion, arrest, and road accidents.
The training set with video-level labels contains 800 normal and 810 abnormal videos, 
while the testing set with frame-level labels has 140 normal and 150 abnormal videos. 

XD-Violence is a multi-modal dataset with videos and audios,
and it includes different backgrounds: movies, games, live scenes, and etc.
There are 4754 videos in the datasets,
which contains 3954 training videos with video-level labels and 800 testing videos with frame-level labels.

\noindent \textbf{Evaluation Metrics.} We evaluate WS-VAD performance with the area under curve (AUC) of the frame-level receiver operating characteristics (ROC) for UCF-Crime.
Meanwhile, the AUC of the frame-level precision-recall curve (AP) is used for XD-Violence.
False alarm rate (FAR) and anomaly subset with only abnormal data are also applied in ablation experiments \cite{CRFD,MIST}.

\subsection{Implementation Details}
We extract the snippet features from pre-trained I3D on Kinetics-400.
The same configurations are set for both datasets.
In the training stage, each video is divided into $N=200$ snippets, and the learning rate is 0.0001.
We train the model with batch size of 64 and iterator time of 3000,
and set the $M^a=60$ and $M^n=60$ for UCF-Crime and XD-Violence.  
The hyperparameters of $(\lambda_1,\lambda_2,\lambda_3,\lambda_4)$ are set as $(0.1, 0.1, 0.001, 0.0001)$ to balance the total loss.
All the operations of topK are equal to $K=\lfloor \frac{M}{16} \rfloor+1$.
In equation (\ref{dis}), $d=100$.

\begin{table}[t] 
    \begin{center}
        \resizebox{0.9\columnwidth}{!}{
            \begin{tabular}{c|ccc}
            \hline
                                        & Method                & Feature(RGB) & AUC(\%)   \\ \hline
            \multirow{3}{*}{Un}         & Hasan et al.(2016)    & N/A          & 50.60      \\
                                        & Lu et al.(2013)       & N/A          & 65.51     \\
                                        & GCL(2022)             & ResNext      & 74.20      \\ \hline
            \multirow{13}{*}{Weakly}    & Sultani et al.(2018)  & C3D          & 75.41     \\
                                        & Sultani et al.(2018)  & I3D          & 76.21     \\
                                        & MA(2019)              & PWC          & 79.00      \\
                                        & GCN (2019)            & TSN          & 82.12    \\
                                        & IBL (2019)            & C3D          & 78.66     \\
                                        & HL-Net (2020)         & I3D          & 82.44    \\
                                        & CLAWS (2020)          & C3D          & 83.03    \\
                                        & MIST (2021)           & I3D          & 82.30     \\
                                        & RTFM (2021)           & I3D          & 84.30     \\
                                        & CRFD (2021)           & I3D          & 84.89     \\
                                        & MSL (2022)            & I3D          & 85.30     \\
                                        & MSL  (2022)           & VideoSwin    & 85.62     \\ \cline{2-4}
                                        & \textbf{Ours}         & I3D          & \textbf{86.97}    \\ \hline
            \end{tabular}
        }    
    \end{center}
    \caption{
    Comparison of the frame-level performance AUC on UCF-Crime testing set with existing approaches.
    }
    \label{table:ucf_res}
\end{table}

\subsection{Results on UCF-Crime}
On UCF-Crime dataset, we compare the AUC scores with existing state-of-the-art (SOTA) VAD methods in Table \ref{table:ucf_res}.
All the methods use RGB features, and our method is based on 10-crop augmentation I3D features.
As the results show, without the prior information of anomaly, unsupervised methods are generally lower in rank than weakly supervised methods.  
When compared with weakly supervised approaches, our method outperforms all the previous methods and achieves the SOTA AUC score of 86.97\%.
There is a 1.35\% improvement for our UR-DMU model when compared to the previous SOTA method MSL (85.62\% AUC)~\cite{MSL},
which is a Transformer-based method with self-training. 
Compared with the MIL-based method RTFM (84.30\% AUC) \cite{RTFM}, we achieve a considerable improvement of 2.67\%.

\subsection{Results on XD-Violence}
The AP scores in XD-Violence are displayed with SOTA VAD methods in Table \ref{table:xd_res}.
We use the 5-crop I3D RGB and features and VGGish audio features provided by \cite{XD} for experiments,
and achieve 81.66\% performance with only RGB features.
Compared with the MIL-based method RTFM (77.81\% AP), we achieve a 3.85\% improvement.
Multi-modal methods are also showed in Table \ref{table:xd_res}. 
With only RGB features, our method outperforms the ACF \cite{ACF}, which utilizes the attention module to fuse the video and audio modal.
We also obtain the same performances as \cite{Pang}.
To evaluate the robustness in multi-modal, we utilize feature concatenation to fuse video and audio features, 
We can also achieve the best performances with an 81.77\% AP score. 
The results of both datasets indicate our UR-DMU is effective in detecting abnormal events with disturbances.

\begin{table}[t]
    \small
    \begin{center}
        \resizebox{0.85\columnwidth}{!}{
            \begin{tabular}{c|ccc}
                \hline
                Method               & Feature        & AP(\%) \\ \hline
                Hasan et al.(2016)   & RGB            & 30.77  \\
                Sultani et al.(2018) & RGB            & 73.20   \\
                HL-Net (2020)        & RGB            & 73.67  \\
                HL-Net (2020)        & RGB+Audio      & 78.64  \\
                CRFD (2021)          & RGB            & 75.90   \\
                RTFM(2021)           & RGB            & 77.81  \\
                MSL(2022)            & RGB            & 78.28  \\
                MSL(2022)            & RGB            & 78.59  \\
                Pang et al.(2021)    & RGB+Audio      & 81.69  \\ 
                ACF(2022)            & RGB+Audio      & 80.13  \\ \hline
                \textbf{Ours}        & RGB            & \textbf{81.66}  \\
                \textbf{Ours}        & RGB+Audio      & \textbf{81.77}  \\ \hline
            \end{tabular}
        }      
    \end{center}
    \caption{
    Comparison of frame-level performance AP on XD-Violence validation set. 
    }
    \label{table:xd_res}
\end{table}
\begin{table}[!t]
    
    \begin{center}
        \resizebox{0.95\columnwidth}{!}{
            \begin{tabular}{cll|cccccc}
            \hline
            \multicolumn{3}{c|}{\multirow{3}{*}{Memory}} & \multicolumn{6}{c}{Dataset}                                            \\ \cline{4-9} 
            \multicolumn{3}{c|}{}                        & \multicolumn{3}{c|}{UCF(\%)}              & \multicolumn{3}{c}{XD(\%)} \\ \cline{4-9} 
            \multicolumn{3}{c|}{}                        & AUC                  & AP            & \multicolumn{1}{c|}{FAR}          & AUC               & AP                & FAR    \\ \hline
            \multicolumn{3}{c|}{Single A-Mem}            & 83.72                & 31.46         & \multicolumn{1}{c|}{1.32}         & 88.07             & 77.59             & 1.38   \\
            \multicolumn{3}{c|}{Single N-Mem}            & 85.19                & 35.23         & \multicolumn{1}{c|}{\textbf{0.98}}& 93.69             & 80.40             & \textbf{0.48}   \\
            \multicolumn{3}{c|}{Dual Mems}               & \textbf{86.97}       & \textbf{35.59}& \multicolumn{1}{c|}{1.05}         & \textbf{94.02}    & \textbf{81.66}    & 0.65  \\ \hline
            \end{tabular}
            }      
    \end{center}
    \caption{
    Comparison of the results of different memory units.
    }
    \label{table:ab_mm}
\end{table}

\begin{table}[!t]
    
    \begin{center}
        \resizebox{0.9\columnwidth}{!}{
        \begin{tabular}{cllcccccc}
        \hline
        \multicolumn{3}{c|}{}                         & \multicolumn{6}{c}{Dataset}                                                                                                                                                                                \\ \cline{4-9} 
        \multicolumn{3}{c|}{}                         & \multicolumn{3}{c|}{UCF(\%)}                                                                                   & \multicolumn{3}{c}{XD(\%)}                                                                \\ \cline{4-9} 
        \multicolumn{3}{c|}{\multirow{-3}{*}{($M_A$,$M_N$)}} & AUC                    & AP                           & \multicolumn{1}{c|}{FAR}                         & AUC                          & AP                           & FAR                         \\ \hline
        \multicolumn{3}{c|}{20,20}                     & 86.71                        & 31.28                        & \multicolumn{1}{c|}{0.82}                        & 93.63                        & 81.13                        & 0.39                        \\
        \multicolumn{3}{c|}{30,30}                     & 86.44                        & 31.29                        & \multicolumn{1}{c|}{1.18}                        & 93.67                        & 80.46                        & 1.18                        \\
        \multicolumn{3}{c|}{40,40}                     & 85.09                        & 32.40                        & \multicolumn{1}{c|}{1.53}                        & 93.07                        & 79.90                        & 0.67                        \\
        \multicolumn{3}{c|}{50,50}                     & 86.28                        & 33.23                        & \multicolumn{1}{c|}{1.63}                        & 93.51                        & 80.19                        & 0.74                        \\
        \multicolumn{3}{c|}{\textbf{60,60}}            & \textbf{86.97}               & \textbf{35.59}               & \multicolumn{1}{c|}{1.05}                        & \textbf{94.02}               & \textbf{81.66}               & 0.65                        \\
        \multicolumn{3}{c|}{70,70}                     & 86.47                        & 33.60                        & \multicolumn{1}{c|}{\textbf{0.49}}               & 93.06                        & 78.87                        & 1.00                        \\
        \multicolumn{3}{c|}{80,80}                     & 85.16                        & 34.18                        & \multicolumn{1}{c|}{0.96}                        & 93.45                        & 79.59                        & \textbf{0.31}               \\
        \multicolumn{3}{c|}{90,90}                     & 86.18                        & 35.14                        & \multicolumn{1}{c|}{2.01}                        & 93.47                        & 79.87                        & 0.44                        \\
        \multicolumn{3}{c|}{100,100}                   & 85.23                        & 31.57                        & \multicolumn{1}{c|}{1.63}                        & 93.92                        & 80.75                        & 0.78                        \\ \hline
        \end{tabular}
    }      
    \end{center}
    \caption{
    Comparison of the results of different memory numbers $M_A$, and $M_N$. $M_A$ is the number of abnormal memory banks, and $M_N$ is the number of normal memory banks.
    }
    \label{table:ab_mn}
\end{table}

\begin{table}[!t]
    \begin{center}
        \resizebox{0.95\columnwidth}{!}{
        \begin{tabular}{cccc|cc}
        \hline
        \multicolumn{4}{c|}{Loss term} & \multicolumn{2}{c}{Dataset} \\ \hline
        $L_{dm}$        & $L_{trip}$        & $L_{kl}$         & $L_{dis}$      & UCF(AUC\%)        & XD(AP\%)    \\ \hline
        \checkmark      & \checkmark        &                  &                & 83.46             & 77.14       \\
        \checkmark      &                   & \checkmark       &                & 82.78             & 78.05       \\
        \checkmark      &                   &                  & \checkmark     & 81.24              & 76.49       \\
                        & \checkmark        & \checkmark       & \checkmark     & 82.09             & 78.74       \\
        \checkmark      &                   & \checkmark       & \checkmark     & 82.78             & 78.16       \\
        \checkmark      & \checkmark        &                  & \checkmark     & 83.92             & 77.57       \\
        \checkmark      & \checkmark        & \checkmark       &                & 86.60             & 80.79       \\
        \checkmark      & \checkmark        & \checkmark       & \checkmark     & \textbf{86.97}    & \textbf{81.66}       \\ \hline
        \end{tabular}
    }
    \end{center}
    \caption{
    Comparison of the results of different loss terms.
    }
    \label{table:ab_loss}
\end{table}

\subsection{Ablation Study}
\textbf{Effectiveness of dual memory units.} 
To verify the effect of the different memory units, we split dual memory units into two single memory units.
As shown in Table \ref{table:ab_mm},
the single abnormal memory unit lacks normal patterns to discriminate what is normality
and receives a higher FAR in normality detection.
It also indicates that abnormal patterns are not easy to learn on weakly supervised data.
The single normal memory unit structure can obtain better results than abnormal memory with rich normal data.
The dual memory units get the best performances because of storing both memory patterns.
To explore a proper number of the memory banks, we set 9 memory cases to search for the best configuration of memory bank numbers.
The results are shown in Table \ref{table:ab_mn}. 
$M_a$ and $M_n$ mean the number of abnormal and normal memories.
When $M_a=60,M_n=60$, our method achieves the best performances on both datasets.

\noindent \textbf{Effectiveness of different loss terms.}
To explore the impact of four loss terms on our complete model, we report the results of 8 loss combinations in Table \ref{table:ab_loss}.
The results of the first three cases show that the $L_{trip}$ is most effective to separate the distance between normality and anomaly.
From the last five cases, the loss terms of $\{L_{dm}, L_{trip}, L_{kl}\}$ are essential and indispensable for our model.
The $L_{dis}$ has a slight effect on detection performance, and control the border between normal and abnormal.
The above observations reveal that our loss terms are more suitable for working together.

\begin{table*}[t]
    
    \begin{center}
        \resizebox{.85\textwidth}{!}{
    \begin{tabular}{ccc|cccccccccc}
    \hline
    \multicolumn{3}{c|}{\multirow{2}{*}{Module}} & \multicolumn{10}{c}{Dataset}                                                                               \\ \cline{4-13} 
    \multicolumn{3}{c|}{}                        & \multicolumn{5}{c|}{UCF(\%)}                                   & \multicolumn{5}{c}{XD(\%)}                \\ \hline
    GL-MHSA      & DMU         & NUL             & AUC   & AP    & AUC$_{sub}$ & AP$_{sub}$  & \multicolumn{1}{c|}{FAR}  & AUC   & AP    & AUC$_{sub}$  & AP$_{sub}$  & FAR  \\ \hline
    \checkmark   &             &                 & 82.89 & 27.07 & 60.18    & 28.30   & \multicolumn{1}{c|}{4.86} & 91.57 & 75.67 & 76.17    & 76.99   & 0.96 \\
    \checkmark   & \checkmark  &                 & 85.53 & 31.21 & 70.13    & 33.79   & \multicolumn{1}{c|}{1.53} & 93.38 & 79.80 & 81.06    & 81.03   & \textbf{0.62} \\
    \checkmark   &             & \checkmark      & 85.19 & 33.61 & 69.74    & 35.79   & \multicolumn{1}{c|}{2.37} & 93.34 & 78.71 & 78.69    & 79.44   & 0.85 \\
    \checkmark   & \checkmark  & \checkmark      & \textbf{86.97} & \textbf{35.59} & \textbf{70.81}    & \textbf{37.25}   & \multicolumn{1}{c|}{\textbf{1.05}} & \textbf{94.02} & \textbf{81.66} & \textbf{82.36}    & \textbf{82.85}   & 0.65 \\ \hline
    \end{tabular}
    }      
    \end{center}
    \caption{
    Comparison of each module's performance on UCF-Crime and XD-Violence. 
    AUC$_{sub}$ and AP$_{sub}$ are the scores only using abnormal data. 
    FAR is the false alarm rate with only normal data.
    }
    \label{table:ab_md}
\end{table*}

\begin{figure*}[!t]
    \centering
    \includegraphics[width=0.9\textwidth]{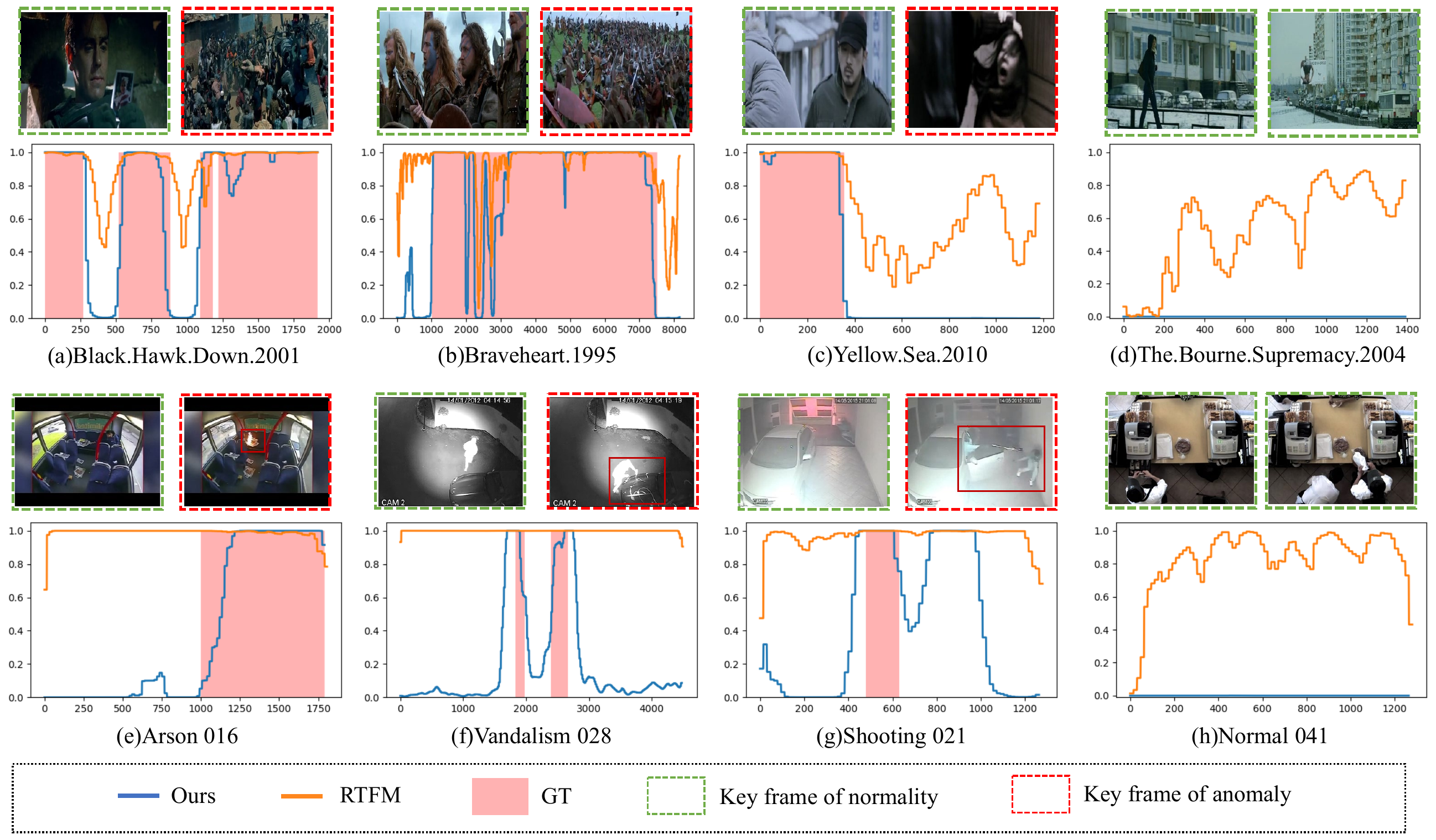} 
    \caption{Qualitative results of anomaly detection performances on UCF-Crime and XD-Violence. 
    The visualizations of (a)$\thicksim $(d) are from XD-Violence, and (e)$\thicksim $(h) are from UCF-Crime.
    }
    \label{fig4}
\end{figure*}

\noindent \textbf{Effectiveness of proposed modules.} 
To investigate the influences of different modules,
we report the results of four combinations of three modules in Table \ref{table:ab_md}, and adopt three additional indicators to comprehensively evaluate our method.
The AUC$_{sub}$ and AP$_{sub}$ are separately the AUC score and AP score of the abnormal video subset, which only contains abnormal videos.
They are used to confirm the anomaly detection performance for anomaly video.
Meanwhile, the FAR is also applied for normal video detection and ensures the performance of normality detection.
GL-MHSA module is the feature learning backbone and achieves the lowest performance.
When adding the DMU module, all the metrics exhibit great improvements. 
The increase of AUC/AP scores and the decrease of FAR indicate that the DMU module can store anomaly and normality patterns.
When NUL module is added, each metric has large improvements.
But it can not map all the normal patterns into Gaussian space without memory banks auxiliary.
Finally, the full model achieves the best performances on two datasets.

\subsection{Qualitative Results}
To further evaluate our method, the qualitative results are shown on two datasets in Figure \ref{fig4}.
We compare our method (blue line) with RTFM (orange line) on eight videos to exhibit the frame-level abnormal predictions.
The pink area is the ground truth area of the anomaly.
We also show the keyframes and mark the abnormal/normal frames to describe what happened in the video, the green rectangles indicate the normal frame and the red rectangles mean the abnormal frame. 
As exhibited in the figure, our method not only produces a precise detection area but also reduces the FAR on the normal part. 
RTFM is based on feature magnitude and is effective for scenes that are smooth and change slowly.
Results of Figure \ref{fig4} (f) and (g) show that RTFM fails to make correct detection.
Our method achieves a lower FAR for normal video (d) and (h). 

\section{Conclusion}
In this work, we propose a UR-DMU model to efficiently separate normality and abnormality for WS-VAD. 
To obtain more expressive embeddings, we introduce the GL-MHSA module to capture long and short temporal dependencies.
Then, we present DMU to store abnormal and normal prototypes under the video-level label,
and the DMU loss is devised to encourage memory units learning and separate them.
Meanwhile, the DUL scheme is proposed to learn latent normal space from rich normal data with noise perturbation.
The predicted anomaly windows and scores of our approach are still relatively coarse and unstable because of using snippet-level features.
In the future, we plan to explore more fine-grained locations in both temporal windows and image space with contrastive learning.

\section{Acknowledgment}
This work is supported by the National Key Research and Development Program of China under Grant
No.2020YFB1805601, National Natural Science Foundation of China (NSFC No. 62272184), and CCF-Tencent Open Research Fund (CCF-Tencent RAGR20220120). The computation is completed in the HPC Platform of Huazhong University of Science and Technology.

{
    \bibliography{aaai23}
}

\end{document}